# A Fingerprint Detection Method by Fingerprint Ridge Orientation Check

Kim JuSong[1], Ri IlYong[1]

(1. Pyongyang Software Joint Development Center, Pyongyang, DPR of Korea)

## Abstract

Fingerprints are popular among the biometric based systems due to ease of acquisition, uniqueness and availability. Nowadays it is used in smart phone security, digital payment and digital locker.

Fingerprint recognition technology has been studied for a long time, and its recognition rate has recently risen to a high level. In particular, with the introduction of Deep Neural Network technologies, the recognition rate that could not be reached before was reached.

In this paper, we propose a fingerprint detection algorithm used in a fingerprint recognition system.

## Keywords

fingerprint; ridge; detection;

## Introduction

The fingerprint recognition system consists of the following steps.

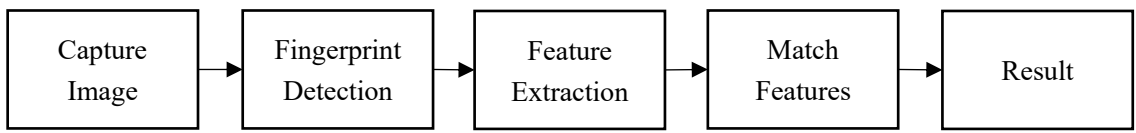

Figure 1. Fingerprint recognition system

In the <Capture Image> step, collect an image using a capture device (optical type, touch type).

In the <Fingerprint Detection> step, check the captured image whether a fingerprint is included.

In the <Feature Extraction> step, extract feature points from the fingerprint image.

In the <Match Features> step, match the extracted feature with the features in the database.

In the <Result> step, return the feature match result.

In this paper, we would like to introduce the algorithm used in the first of these steps, fingerprint detection.

There are many studies for fingerprint detection algorithm. The most frequently used methods include <Fingerprint Detection Method by Image Segmentation and Brightness Value Difference> (Method 1), <Fingerprint Detection Method by Brightness Value Difference> (Method 2), and <Fingerprint Detection Method by Histogram Analysis> (Method 3). These methods are most commonly used in the field of fingerprint recognition. However, these methods work well in optical sensor and touch sensor with good image quality, but they did not perform well in images that mixed a lot of noise when collecting as shown in the figure below.

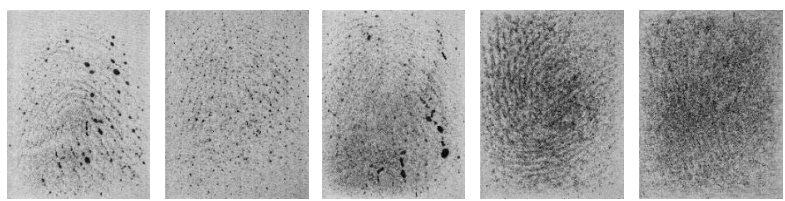

Figure 2 – 1 ~ 5. Noisy image without fingerprint

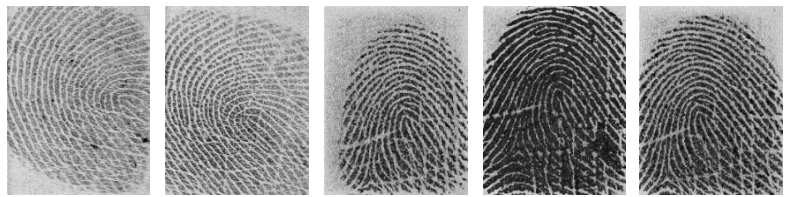

Figure 2 – 6 ~ 10. Normal fingerprint image

In the case of a sensor where noise is mixed in this way, the noise does not disappear and remains even after being released by sweat. Previous methods mistakenly recognize these images as an image that contain fingerprint.

In order to solve this problem, this paper proposed a fingerprint detection algorithm by fingerprint ridge orientation check.

Noisy images without fingerprint, as shown in <Figure 2 – 1 ~ 5>, cannot be correctly detected by general brightness check and area calculation.

Therefore, we decided to calculate the orientation of the fingerprint ridge and determine the presence or absence of a fingerprint.

The fingerprint ridge orientation $\theta_{xy}$ is an angle formed by the fingerprint ridge with the x-axis in the range of the pixel $(x, y)$.

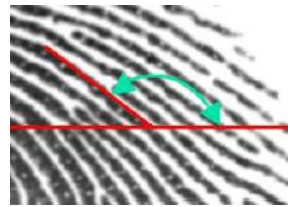

Figure 3. Fingerprint Ridge Orientation

At this time, the checking of the fingerprint ridge orientation was performed by referring to a method used in the <Feature Extraction> step of the fingerprint recognition system.

In the <Feature Extraction> step of the fingerprint recognition system described in the literature [1] and [2], the orientation of the fingerprint ridge is estimated by the following method.

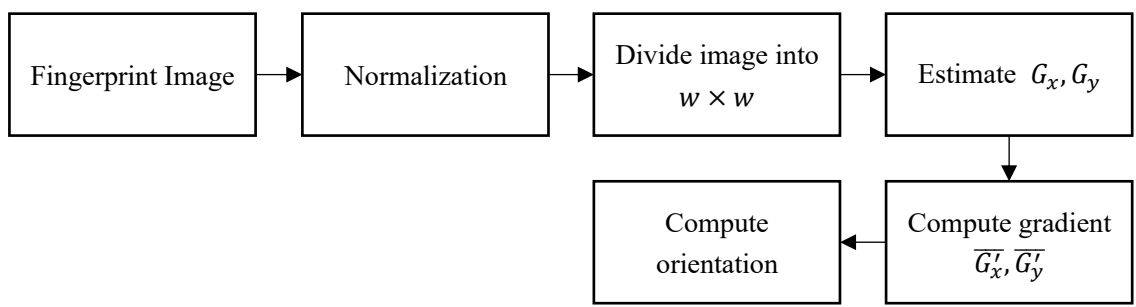

Figure 4. Flow chart for orientation extraction described in literatures

The specific steps are as follows.

First, normalize the image. Alternatively, histogram equalization can also be used.

And then, divide the image into $w \times w$ overlapping blocks.

Next, the gradients $G_x, G_y$ are computed using Sobel filter. $G_x$ represents the horizontal gradient component and $G_y$ represents the vertical gradient component.

The average gradient vectors $\overline{G_x'}, \overline{G_y'}$ are computed as follows:

$$\overline{G'_x} = \frac{1}{w \times w} \sum_{i=1}^{w} \sum_{j=1}^{w} \left(G_x^{\,2}(i,j) - G_y^{\,2}(i,j)\right)$$

$$\overline{G'_y} = \frac{1}{w \times w} \sum_{i=1}^{w} \sum_{j=1}^{w} \left(2G_x(i,j)G_y(i,j)\right)$$

Finally, estimate direction using following formula.

$$\theta = \begin{cases} \frac{1}{2}\tan^{-1}\frac{\overline{G'_y}}{\overline{G'_x}} + \frac{\pi}{2}, & \tan^{-1}\frac{\overline{G'_y}}{\overline{G'_x}} < 0 \\ \frac{1}{2}\tan^{-1}\frac{\overline{G'_y}}{\overline{G'_x}} - \frac{\pi}{2}, & \tan^{-1}\frac{\overline{G'_y}}{\overline{G'_x}} \geq 0 \end{cases}$$

The orientation of the fingerprint ridge is calculated through the above steps.

However, if this method is applied as it is, it will consume a lot of time due to the large amount of calculation.

Unlike the <Feature Extraction> step, in the <Fingerprint Detection> step, the presence or absence of a fingerprint must be detected in real time. In particular, since it must operate in an embedded system with low performance, real-time can be guaranteed by reducing the amount of calculation.

So, in the above steps, we modified some computational steps.

First, in the preprocessing stage of the input image, we applied a simple low-pass filter instead of normalization with a large amount of calculation. In this way, it was possible to increase the speed while removing the high-frequency noise in the image.

Next, we did binarization using the Otsu algorithm, and selected the type of the ridge by the CN (crossing numbering) calculation method. The CN calculation was carried out using the following formula.

| $P_4$ | $P_3$ | $P_2$ |
|---|---|---|
| $P_5$ | $P$ | $P_1$ |
| $P_6$ | $P_7$ | $P_8$ |

Figure 5. Neighborhood of the pixel in 3x3 window

$$CN = 0.5 \sum_{i=1}^{8} |P_i - P_{i+1}|$$

And then calculate the Gradient $G_x, G_y$ only for the ridge lines having CN values of 1(Ridge ending point), 3(Bifurcation point), 4(Crossing point). At this time, gradient was calculated using Sobel filter.

Next, calculate the average gradient $\overline{G'_x}, \overline{G'_y}$ by setting the window size $w$ to 16.

At this time, when $\overline{G'_x}, \overline{G'_y}$ are not 0, the direction is considered to be present, and increase the number of feature points.

After calculating the total image in this way, if the number of feature points exceeds the threshold at the end, it was determined as an image with a fingerprint, and if not, it was determined as an image without a fingerprint.

The threshold value was selected in an experimental method, it was selected as a value between 150 and 200 for 256 x 360 images.

In addition, in order to increase the calculation speed, only 2/3 of the area was processed based on the center without processing the entire image. Also, when the image size is large, the binarization was not performed using the Otsu algorithm, but the general threshold values (100 to 150) were used.

In this way, in this paper, we proposed one method of performing fingerprint detection by calculating the orientation of the fingerprint ridge.

## Results and discussion

As a result of performance evaluation, images of <Figure 2 – 1 ~ 5>, which were not properly detected in the representative methods used by various fingerprint modules, were also accurately detected.

Here we showed the results of testing the images of <Figure 2 – 1 ~ 10> using various methods.

| Method | Image1 | Image2 | Image3 | Image4 | Image5 | Image6 | Image7 | Image8 | Image9 | Image10 |
|---|---|---|---|---|---|---|---|---|---|---|
| **Proposal** | **OK** | **OK** | **OK** | **OK** | **OK** | **OK** | **OK** | **OK** | **OK** | **OK** |
| Method 1 | Fail | Fail | Fail | Fail | Fail | Fail | OK | OK | OK | OK |
| Method 2 | OK | Fail | Fail | Fail | Fail | Fail | OK | OK | OK | OK |
| Method 3 | OK | Fail | Fail | Fail | Fail | OK | OK | OK | OK | OK |
| Method 4 | OK | OK | OK | Fail | Fail | OK | OK | OK | OK | OK |
| Method 5 | OK | OK | OK | Fail | Fail | OK | OK | OK | OK | OK |

Here is a description of the various methods and the calculation quantity. (Measured at 120MHz MCU)

| Method | Description | Elapsed Time (ms) |
|---|---|---|
| **Proposal** | **Fingerprint Detection by Fingerprint Ridge Orientation Check** | **4** |
| Method 1 | Fingerprint Detection by Image Segmentation and Brightness Value Difference | 1 |
| Method 2 | Fingerprint Detection by Brightness Value Difference | 1 |
| Method 3 | Fingerprint Detection by Histogram Analysis | 2 |
| Method 4 | Fingerprint Detection by Difference in Brightness Values in a Specific Zone | 3 |
| Method 5 | Fingerprint Detection by Reverse Area of Fingerprint Command | 5 |

In this way, it was possible to sufficiently guarantee real-time speed in an embedded system while maximizing performance by conducting fingerprint detection by fingerprint ridge orientation check.

In addition, as a result of experimenting with 800 fingerprint images in the database FVC2004 DB1_A, the fingerprint state was detected 100% accurately by the method proposed by this paper. From this, it was confirmed that the method proposed in this paper works accurately not only for noise-mixed images but also for normal images.